\documentclass{article}

\usepackage{microtype}
\usepackage{forest}
\usepackage{xcolor}
\usepackage{graphicx}
\usepackage{subfigure}
\usepackage{booktabs} 
\usepackage{amsmath}
\usepackage{graphicx} 
\usepackage{booktabs} 
\usepackage{array}
\usepackage{listings}
\usepackage{xcolor}
\definecolor{mygreen}{RGB}{0,150,0}
\usepackage{hyperref}

\usepackage[accepted]{icml2024}

\usepackage{amsmath}
\usepackage{amssymb}
\usepackage{mathtools}
\usepackage{amsthm}

\usepackage[capitalize,noabbrev]{cleveref}

\theoremstyle{plain}
\newtheorem{theorem}{Theorem}[section]
\newtheorem{proposition}[theorem]{Proposition}

\theoremstyle{definition}

\theoremstyle{remark}

\usepackage[textsize=tiny]{todonotes}

\icmltitlerunning{Injecting Hierarchical Biological Priors into GNNs for Flow Cytometry Prediction}

\begin{document}

\twocolumn[
\icmltitle{Injecting Hierarchical Biological Priors into Graph Neural Networks for Flow Cytometry Prediction}



\icmlsetsymbol{equal}{*}

\begin{icmlauthorlist}
\icmlauthor{Fatemeh Nassajian Mojarrad}{equal,yyy}
\icmlauthor{Lorenzo Bini}{equal,yyy}
\icmlauthor{Thomas Matthes}{comp}
\icmlauthor{Stéphane Marchand-Maillet}{yyy}
\end{icmlauthorlist}

\icmlaffiliation{yyy}{Department of Computer Science, University of Geneva, Switzerland}
\icmlaffiliation{comp}{Hematology Service, Department of Oncology and Clinical Pathology Service, Geneva University Hospital, Switzerland}

\icmlcorrespondingauthor{Fatemeh Nassajian Mojarrad}{Fatemeh.Nassajian@unige.ch}
\icmlcorrespondingauthor{Lorenzo Bini}{Lorenzo.Bini@unige.ch}
\icmlcorrespondingauthor{Thomas Matthes}{Thomas.Matthes@hcuge.ch}
\icmlcorrespondingauthor{Stéphane Marchand-Maillet}{Stéphane.Marchand-Maillet@unige.ch}

\icmlkeywords{Machine Learning, ICML, Graph Neural Networks, Prior Knowledge, Biology}

\vskip 0.3in
]



\printAffiliationsAndNotice{\icmlEqualContribution}  

\begin{abstract}
In the complex landscape of hematologic samples such as peripheral blood or bone marrow derived from
flow cytometry (FC) data, cell-level prediction presents profound challenges. This work explores injecting hierarchical prior knowledge into graph neural networks (GNNs) for single-cell multi-class classification of tabular cellular data. By representing the data as graphs and encoding hierarchical relationships between classes, we propose our hierarchical plug-in method to be applied to several GNN models, namely, FCHC-GNN, and effectively designed to capture neighborhood information crucial for single-cell FC domain. Extensive experiments on our cohort of 19 distinct patients, demonstrate that incorporating hierarchical biological constraints boosts performance significantly across multiple metrics compared to baseline GNNs without such priors. The proposed approach highlights the importance of structured inductive biases for gaining improved generalization in complex biological prediction tasks.
\end{abstract}

\section{Introduction}
\label{Intro}
Flow cytometry (FC) is a technology that provides rapid multi-parametric analysis of single cells in solution. The expression of specific cell surface and intracellular molecules is thereby detected by antibodies coupled to fluorochromes. Exposing labeled cells to a laser leads to the emission of fluorescence which is then collected by a detector. By analyzing more than millions of cells in a short time, information can thereby be collected about the presence or absence of up to fifty different cellular molecules per cell. As a result, histologists obtain tabular data where each row corresponds to a single cell and every column to a marker or feature (cell surface molecule -- Table \ref{tab:1}, in Appendix \ref{appendix:data-description}). 

This technique is employed in medicine to assess the cellular composition of complex body fluids like blood or bone marrow and to facilitate the diagnosis of hematologic diseases such as leukemia. 
In collaboration with our University's Hospital, we have built a dataset obtained by the analysis of bone marrow samples from 19 healthy/recovered patients with information obtained up to $1'512'610$ cells/sample about the presence or absence of twelve different surface molecules. Cellular processes are governed by complex hierarchical relationships and neighborhood interactions that are challenging to capture using traditional flat data representations. This strongly motivates our decision to apply GNNs, which have emerged as a powerful class of models capable of leveraging graph-structured data to reason about relationships and dependencies. By representing biological entities and their associations as nodes and edges in a graph, GNNs can effectively encode neighborhood information and propagate signals across the graph topology during training. However, even with this graph-based inductive bias, GNN models may still struggle to generalize well without additional guided constraints from biological domain knowledge.

In this work, we propose a novel approach to inject hierarchically-structured priors (biological in our case) into the GNN framework for multi-class prediction tasks on tabular data. Specifically, we encode the known hierarchical relationships between different cell types or functional classes as a tree-structured hierarchy imposed on the GNN output space. This hierarchical constraint serves as an inductive bias that encourages the GNN to respect the hierarchical dependencies inherent to the biological domain.
During training, we use a custom hierarchical loss function that accounts for the hierarchical similarities between classes, in addition to the traditional cross-entropy loss. This approach ensures that the model's predictions not only accurately classify instances into their respective leaf nodes (specific cell types), but also respect the hierarchical groupings at higher levels of the taxonomy (broader cell lineages or functional categories).
Through rigorous experiments on our real-world datasets, we demonstrate that incorporating hierarchical biological priors significantly boosts the performance of GNN models across a range of metrics. Our hierarchical GNN approach achieves consistent gains over strong baselines that do not leverage such structured domain knowledge. We provide in-depth analysis and empirical evidence highlighting the importance of encoding inductive biases aligned with the underlying data domains, especially for domains with rich hierarchical relationships like cell biology.

\section{Related Work}
\label{Related Work}
The primary research focus in machine learning has predominantly been centered on developing models for conventional classification problems, wherein an object is assigned to a single class from a set of disjoint classes. However, a distinct subset of tasks involves scenarios where classes are not disjoint but rather organized hierarchically, giving rise to hierarchical classification (HC). In HC, objects are linked to a specific superclass along with its corresponding subclasses. Depending on the nature of the task, this association may encompass all subclasses or only a designated subset. 
The hierarchical structure formalizing the interrelation among classes can manifest as either a tree or a directed acyclic graph (DAG) \cite{Silla2011}. Hierarchical classification problems manifest across a broad spectrum, spanning from musical genre classification \cite{Ariyaratne2012,Iloga2018} to the identification of COVID-19 in chest X-ray images \cite{Pereira2020}, the taxonomic classification of viral sequences in metagenomic data
\cite{Shang2021} and text categorization \cite{Javed2021,Ma2022}. 
Graph-based models can leverage both the global and local characteristics inherent in networks relevant to cell biology. GNNs have been recently adapted in various
tasks, e.g. link prediction \cite{Zhang2018}, graph classification \cite{Duvenaud2015,Lee2019}, node classification \cite{Kipf2017,Velickovic2017,Yang2016}, and single-cell RNA sequencing \cite{vandijk2020}. For FC data FlowCyt \cite{bini:chil2024} represents the first publicly available benchmark, designed to test deep learning models for single-cell multi-class classification, highlighting the benefits of modeling these data as a graph-structured problem. 
In the context of HC problems, DiffPool \cite{ying2018hierarchical} propose a differentiable graph pooling module to generate hierarchical representations of graphs using various GNNs in an end-to-end fashion. Moreover, in the biological domains, HC-GNNs have been widely applied for histological image classification \cite{hou2022spatial}, protein-protein interaction \cite{gao2023hierarchical}, molecules' properties prediction \cite{han2023himgnn} and for cancer diagnosis and prognosis in digital pathology \cite{pati2022hierarchical}.

\section{Problem Description}
\label{Problem Description}
\begin{figure}[ht]
\vskip 0.2in
\begin{center}
\centerline{\scalebox{0.8}{\begin{forest}
  for tree={
    edge={->, thick},
    l=1.5cm
  }
  [Total cell population
    [\textcolor{blue}{CD45 pos}
      [\textcolor{blue}{Lymphocytes}
        [\textcolor{blue}{B cells}
          [\textcolor{mygreen}{Lambda pos}]
          [\textcolor{mygreen}{Kappa pos}]
        ]
        [\textcolor{mygreen}{NK cells}]
        [\textcolor{blue}{T cells}
          [\textcolor{mygreen}{CD8 T cells}]
          [\textcolor{mygreen}{CD4 T cells}]
        ]
      ]
      [\textcolor{mygreen}{Monocytes}]
      [\textcolor{mygreen}{Neutrophils}]
    ]
    [\textcolor{mygreen}{CD45 neg}]
  ]
\end{forest}}}
\caption{Depiction of our HC set up.}
\label{fig:tree1}
\end{center}
\vskip -0.2in
\end{figure}
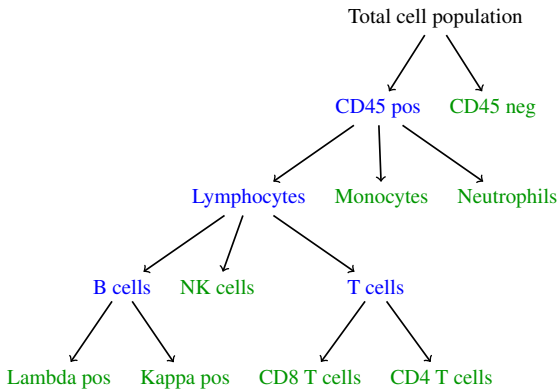

In this research, we began by collecting raw data from the bone marrow of 19 patients who underwent a flow cytometric analysis for diagnostic purposes. All samples were processed by the Diagnostics laboratory of the University Hospital and no malignant disease was detected at this stage. Each patient's sample is primarily tabular data (see supplementary material for the futher details) composed by a $N\times D$ matrix of values, where $D = 12$ dimensions (see Table \ref{tab:1} in Appendix \ref{appendix:data-description} for details). It is therefore equivalent to a $D$-dimensional point cloud that cytometrists visualize via 2D projections. The result of such operations is the definition and quantification of different cell populations according to their phenotype. Those populations of interest were defined according to the markers expression, and then grouped together into different categories as shown in Figure \ref{fig:tree1}, serving as our main biological hierarchical prior in this study. 

The main focus is to classify different cell types present in the heterogeneous samples into 12 classes, where 8 of them belong to leaf nodes (\textcolor{mygreen}{green ones}), while 4 of them are parental nodes (\textcolor{blue}{blue ones}).

\section{Archithecture of FCHC-GNN}
\label{sec:FCHC-GNN}
Given a tabular dataset containing $n$ data samples and $m$ feature fields, denoted by $X=[\mathbf{x}_1, \cdots,\mathbf{x}_n]^T$, the $i$-th sample in the table associated with $m$ feature values $\mathbf{x}_i=[ x^{(1)}_i, \cdots,x^{(m)}_i]$ and a discrete label $y_i$,  part of a label hierarchy. Our learning goal is to find a mapping function $f$ for which given $\mathbf{x}_i$, returns the predicted label $\hat{y}_i$. In the rest of this section, we will describe our plug-in method FCHC-GNN, and the architecture of the network for the underlying classification problem.

We consider an HC problem with a given set $\mathcal{C}$ of $C$ classes, which are hierarchically organized as a directed tree (Figure \ref{fig:tree1} for example). If there exists a path of non-negative length from class $A_1$ to class $A_2$ in the tree, we call $A_2$  a subclass of $A_1$. We assume to have a mapping $f_A: \mathbb{R}^m \to [0,1]$ for every class $A$ such that $\mathbf{x}\in \mathbb{R}^m$ is predicted to belong to $A$ whenever $f_A(\mathbf{x})$ is greater than or equal to some threshold. When $A_2$ is a subclass of $A_1$, the model should guarantee that $f_{A_1}(\mathbf{x})\geq f_{A_2}(\mathbf{x})$, for all $\mathbf{x}\in \mathbb{R}^m$, to guarantee that the hierarchy constraint is always satisfied independently from the threshold. 
Note that in this model, every class is considered a subclass of itself. The case where $f_{A_1}(\mathbf{x})<f_{A_2}(\mathbf{x})$ for some $\mathbf{x}\in \mathbb{R}^m$ and $A_2$ is a subclass of $A_1$, is called a {\em hierarchy violation} \cite{Wehrmann2018}. To simplify the notations, we will write $f_A(\mathbf{x})$ as $f_A$,  thus removing the functional dependency of this term concerning $\mathbf{x}$.

\paragraph{Overall structure} Given a set $\mathcal{C}=\{A_1,\cdots, A_C\}$ of $C$ hierarchically structured classes $A_c$, we first build a neural network composed of two modules:  
a {\em Early Module} $\mathcal{H}$ with one output for each class in $\mathcal{C}$ and an upper module, referred to as the {\em Max Constraint Module} (MCM), consisting of a single layer that takes the output of the early module as input and imposes the hierarchy constraint.

If $y_A$ is the ground truth label for class $A\in \mathcal{C}$ and the prediction value $\mathcal{H}_A \in [0,1]$ is obtained by $\mathcal{H}$ for class $A$, the MCM output for a class $A$ is given by:
\begin{equation}
\label{MCM}
    \text{MCM}_A=\max_{B\in \mathcal{S}_A} \mathcal{H}_B.
\end{equation}
where $\mathcal{S}_A$ is the set of subclasses of $A$ in $\mathcal{C}$. 
Note that since the number of operations performed by $\text{MCM}_A$ is independent of the depth of the hierarchy, for each class $A \in \mathcal{C}$, the resulting model is scalable. 

By construction, the following theorem therefore holds.
\begin{theorem}
\label{theorem1}
Let $\mathbf{x}\in \mathbb{R}^m$ be a data point.
Let $\mathcal{C}=\{A_1,\cdots, A_C\}$  be a set of hierarchically structured classes and let $\mathcal{H}$ be a early module with outputs $\mathcal{H}_{A_1},\cdots, \mathcal{H}_{A_C}$ ($\mathcal{H}_{A_c}\in[0,1]~\forall c$) given the input $\mathbf{x}$, and $\text{MCM}_{A_1},\cdots, \text{MCM}_{A_C}$ be defined as in equation \eqref{MCM}. 
Then FCHC-GNNs do not admit hierarchy violations.
\end{theorem}
Second, to exploit the hierarchy constraint during training, FCHC-GNNs are trained with {\em max constraint loss} (MCLoss), defined as follows
\begin{equation*}
   \text{MCLoss}_A=-y_A \ln (\max_{B\in \mathcal{S}_A} y_B \mathcal{H}_B ) -(1-y_A) \ln (1-\text{MCM}_A),
\end{equation*}
for each class $A\in \mathcal{C}$.
The global MCLoss is written as 
\begin{equation}
  \text{MCLoss}= \sum_{A\in \mathcal{C}}  \text{MCLoss}_A.
\end{equation}
The advantage of defining MCLoss instead of using the standard binary cross-entropy loss is that the more ancestors a class has, the more likely it is that FCHC-GNN trained with the standard binary cross-entropy loss would remain stuck in spurious local optima. The MCLoss prevents this from happening. 

As a result, FCHC-GNN can delegate the prediction for a class $A_i$ to one of its subclasses $A_j$, thanks to MCM and the overall hierarchical constrained loss function MCLoss. In fact, $\text{MCM}_A$ can be viewed as the maximum score among subclasses of $A$ being used within the MCLoss. Therefore, $\mathcal{H}_B$ are the prediction scores for subclasses used in computing both $\text{MCM}_A$ and MCLoss.
More formally, the following proposition holds by construction.

\begin{proposition}
Under the same assumptions as in Theorem~\ref{theorem1}, considering a class $A_i\in \mathcal{C}$ and $A_j\in \mathcal{S}_{A_i}$ with $i \neq j$, FCHC-GNN delegates the prediction on $A_i$ to $A_j$ for $\mathbf{x}$, if $\text{MCM}_{A_i}=\mathcal{H}_{A_j}$ and $\mathcal{H}_{A_i}<\mathcal{H}_{A_j}$.
\end{proposition}
\paragraph{Classification layer}
We have designed the FCHC module to be compatible with GAT, SAGE, and GCN, as highlighted in our code publicly accessible at \url{https://github.com/VIPER-GENEVA/FCHC-GNN-Hierarchical}\footnote{Check out VIPER official page for latest work \url{https://viper-geneva.github.io/}.}, but due to its general nature it can be easily extended to broader GNNs. For the sake of simplicity, in the following discussion, we explain how we built the Early Module $\mathcal{H}$ for the classification as a Graph Attention Network (GAT), following \cite{Velickovic2017}. Given $\{\mathbf{x}_i\}_{i=1}^n$ as nodes, and $\mathbf{x}_i\in \mathbb{R}^m$, the node features are the input of the graph attention layer.  
The output of the layer is a new set of node features $\{\mathcal{H}_i\}_{i=1}^n$, where $\mathcal{H}_i \in \mathbb{R}^{\Tilde{m}}$, for which $\Tilde{m}$ might be different from $m$. 

A GAT defines {\em attention coefficients} as importance scores of node $\mathbf{x}_j$'s features to that of node $\mathbf{x}_i$.
$W$ is the weight matrix used to build attention coefficients.
{\em Masked attention } normalizes {\em attention coefficients} within every graph neighborhood $\mathcal{N}(\mathbf{x}_i)$.
In our case, $\mathcal{N}(\mathbf{x}_i)$ represents the Euclidean $k$-nearest neighbours of node $\mathbf{x}_i$.
The attention mechanism is a single-layer feedforward neural network, parameterized by a weight vector $\vec{a} \in \mathbb{R}^{2\Tilde{m}}$, over which we apply the LeakyReLU non-linearity ( see Fig. \ref{fig:network}(a) in Appendix \ref{appendix:gat-viz}). 

The normalized attention coefficients $\gamma_{ij}$ are then used to compute a linear combination of the corresponding features, after applying a nonlinearity $\sigma$, to serve as the final output features for each node. 

We also move from single-head attention to multi-head attention (with $L$ heads) by concatenating the output of all heads at every layer $i$ (but the final layer).
The expression for the final layer replaces concatenation by averaging and can be written in our notation as
\begin{equation}
 \mathcal{H}_i= \sigma (\frac{1}{L}\sum_{l=1}^L \sum_{j \in \mathcal{N}(\mathbf{x}_i)} \gamma_{ij}^l W^l\mathbf{x}_j).   
\end{equation}

\section{Experiments} 
\label{sec:Experiments}
In this section, we present the experimental results to illustrate the behaviour of ours FCHC-GNNs for the multi-class classification problem on our FC dataset. We also compare the proposed method with the respective flat versions (meaning without the FCHC module attached), to highlight how the injection of hierarchical prior knowledge really boost the performance for FC classification with respect to flat structures. Moreover, since they are known to be highly performant on tabular data, we also incuded Deep Neural Network both with (FCHC-DNN) and without (DNN) hierarchical module. We have also obtained competitive performance of our FCHC-GNN module on the public Imagenet1000-(Mini) dataset, where we used the subcategory of "animals"; these additional results are presented in Appendix \ref{appendix:imagenet} due to space constraints.

Furthermore, in Appendix \ref{appendix:1} as ablation studies, we further validate the generality of our FCHC-GNN module by conducting experiments on other 30 different FC patients with a shallower depth of hierarchy. Across this cohort, we observe that injecting the hierarchical priors becomes necessary as the hierarchy grows deeper and more complex. For shallow hierarchies, the performance gains are modest, but they become substantially larger when dealing with intricate, multi-level hierarchies where using prior knowledge becomes fundamental.

Due to space constraints, see Appendix \ref{appendix:data-description} for experimental setup, metrics definition and graph-construction details. Please see also Appendix \ref{appendix:1} for model's ablation studies on different hierarchy scenario and hyperparameters details.
Our hierarchical approach achieves consistent gains over same architectures that do not leverage such structured domain knowledge, as shown in Table \ref{tab:4}, \ref{tab:resFCHCGNN} and \ref{tab:res-no-FCHCGNN}, where we compute hierarchical-metrics and the correct predicted classes across all the patients in the dataset. All the results are within the $\text{std} = \pm 0.1$, and experiments have been averages across four different seeds.

\begin{table}[ht]
\caption{Average metrics across all patients using different models for FCHC module.}
 \label{tab:4}
\vskip 0.15in
\begin{center}
\begin{tiny} 
\begin{sc}
\begin{tabular}{lcccr}
\toprule 
        Metrics  & FCHC-GAT  
        & FCHC-SAGE & FCHC-GCN & FCHC-DNN \\
       \midrule
        hp  & 0.88 
        & \textbf{0.92} & 0.83 & 0.74\\
        hr  &  \textbf{0.85}
        &0.83 & 0.81 & 0.83\\
        hf  & 0.87 
        & \textbf{0.88} & 0.82 & 0.78\\
\bottomrule
\end{tabular}
\end{sc}
\end{tiny}
\end{center}
\vskip -0.1in
\end{table}

\begin{table}[ht]
\caption{Average ratios of correct predicted classes across all patients using different models for FCHC module.}
    \label{tab:resFCHCGNN}
    \vskip 0.15in
\begin{center}
\begin{tiny}
\begin{sc}
\begin{tabular}{lcccr}
\toprule
        Label  & FCHC-GAT  
        & FCHC-SAGE & FCHC-GCN & FCHC-DNN\\
        \midrule
        CD45 pos  & 100.00 
        & 100.00 & 100.00 & 81.12\\
        CD45 neg  & 100.00 
        & 100.00 & 100.00 & 83.43\\
        Lymphocytes  & 85.21 
        &\textbf{97.91} &82.14&81.06\\
        Monocytes & 90.31 
        &93.42 &\textbf{95.52} &90.07\\
        Neutrophils & 80.15 
        & \textbf{93.38} & 75.26 & 80.84\\
        B Cells & 94.86 
        &97.49& \textbf{98.52} & 85.83\\
        NK Cells & 97.48  
        & 97.50& \textbf{97.98} & 94.01\\
        T Cells & 83.00  
        & \textbf{97.84} & 86.74 & 92.07\\
        Lambda pos & 97.77  
        & \textbf{98.88} & 96.52 & 86.45\\
        Kappa Pos & 96.56  
        & 98.61 & \textbf{99.21} & 98.00\\
        CD8 T Cells & 90.15  
        & \textbf{95.26} & 94.25 & 91.99\\
        CD4 T Cells & 89.81  
        & 91.63 &\textbf{92.07} &90.06\\
\bottomrule
\end{tabular}
\end{sc}
\end{tiny}
\end{center}
\vskip -0.1in
\end{table}

\subsection{Discussion of Our Results}
\label{subsec:disc_res}
The results presented in this paper underscore the novelty and effectiveness of our newly introduced FCHC-GNN framework, in performing node classification tasks on FC dataset. However, the presented approach is rather general and can be easily extend to broader applications of GNN, in case there is the need to leverages the inherent hierarchical structure of the data, a feature that frequently appears in real-word dataset. Indeed, as shown in Table \ref{tab:res-no-FCHCGNN} where the FCHC model has not been applied, injecting hierarchical biological priors allows the models to not overlook any of the \textcolor{mygreen}{leaf} nodes during the prediction. As an example, in Table \ref{tab:resFCHCGNN} and \ref{tab:res-no-FCHCGNN} we clearly see how the performances drops from $91.63\%$ with hierarchical priors to a $67.80\%$ without, on CD4 T cells for GraphSAGE model \cite{hamilton2017inductive}. Similarly, for NK cells the correct predicted labels drammatically goes down from $97.48\%$ to $0.15\%$ for the GAT, highlighting once again the need of including the FCHC module to caputure the biology complexity of these cells population. 
\begin{table}[ht]
\caption{Average ratios of corrected predicted classes across all patients without using the FCHC module.}
    \label{tab:res-no-FCHCGNN}
    \vskip 0.15in
\begin{center}
\begin{scriptsize}
\begin{sc}
\begin{tabular}{lcccr}
\toprule
        Label  & GAT  
        & SAGE & GCN & DNN\\
        \midrule
        CD45 neg  & 34.94  
        & 98.25 &97.10 &\textbf{98.44}\\
        Monocytes  & 0.75  
        & \textbf{1.46} & 0.50 & -\\
        Neutrophils & \textbf{100.00} 
        & 98.96 & 99.73 & 98.77\\
        NK Cells & 0.15  
        & \textbf{1.57} & \textbf{-} & -\\
        Lambda pos & -  
        & 0.88 & 0.07 & \textbf{1.40}\\
        Kappa Pos & 0.24  
        & - & 0.10 & \textbf{0.70}\\
        CD8 T Cells & -  
        & \textbf{20.41} & 13.68 & 1.01\\
        CD4 T Cells & 0.20  
        & \textbf{67.80} & 40.08 &31.95\\
\bottomrule
\end{tabular}
\end{sc}
\end{scriptsize}
\end{center}
\vskip -0.1in
\end{table}

Given the classification power of FCHC module, we took a deeper look at each model's performance, and 
to allow better visualization of the predicted cell patterns, we picked one random patient and we plotted its t-SNE embeddings \cite{t-SNE}, as shown in Figure \ref{fig:tSNE} of Appendix \ref{appendix:tsne}. Moreover, we further investigated the interpretability of our module through the expression of feature importance, as resulted in Appendix \ref{appendix:feat-importance}.  

\section{Conclusion and Future Work}
\label{sec:Conclusion-FutWork}
We have presented a novel hierarchical GNN framework, namely FCHC-GNN, that injects structured biological priors to enhance multi-class prediction on FC cellular data. By encoding known hierarchical relationships between cell types and functional classes, our model effectively captures the rich dependencies innate to biological domains while operating on a tabular input representation. Extensive experiments across our FC dataset demonstrate substantially improved performance over strong baselines that fail to leverage such hierarchical constraints. Moreover, as demonstrated over well-known ImageNet data, our FCHC-GNN module is easily extendible to other domains where classes may be organized within a hierarchy.

The consistent gains validate our hypothesis that tailored inductive biases aligned with the underlying tabular data distributions are crucial for achieving better generalization.
FCHC-GNN architectures introduce a novel method to enhance learning on tabular inputs with prior domain knowledge expressed as hierarchies. Looking ahead, this general approach holds promise for biomedical learning tasks beyond single-cell classification.
\section*{Acknowledgements}
This work is partly funded by the Swiss National Science Foundation under grant number 207509 "Structural Intrinsic Dimensionality".

\bibliography{main}
\bibliographystyle{icml2024}

\newpage
\appendix
\onecolumn
\section{Description of Our Dataset}
The raw data obtained from the cytometer are saved in Flow Cytometry Standard (FCS) files. Each FCS file comprises multidimensional data that corresponds to more than one million of cells, with each cell characterized by various parameters such as size, granularity, and fluorescence intensity, giving them the structure of tabular data (see Table \ref{tab:1} for a complete list of markers in our data).
\label{appendix:data-description}
\begin{table}[ht]
\caption{Flow cytometry data markers.}
\label{tab:1}
\vskip 0.15in
\begin{center}
\begin{scriptsize}
\begin{sc}
\begin{tabular}{lcccr}
\toprule
 Marker \# & Marker  & Explanation  \\
\midrule
       0&  FS INT & Forward Scatter (FSC) - Cell’s size \\
       1& SS INT & Side Scatter (SSC) - Cell’s granularity\\
       2&  CD14-FITC & Cluster of Differentiation 14 - Antigen\\
       3& CD19-PE & Cluster of Differentiation 19 - Antigen\\
       4& CD13-ECD &Cluster of Differentiation 13 - Antigen \\
      5&  CD33-PC5.5 & Cluster of Differentiation 33 - Antigen\\
      6&  CD34-PC7 & Cluster of Differentiation 34 - Antigen\\
      7&  CD117-APC &Cluster of Differentiation 117 - Antigen \\
       8& CD7-APC700 &Cluster of Differentiation 7 - Antigen \\
      9&  CD16-APC750 &Cluster of Differentiation 16 - Antigen \\
      10&  HLA-PB &Human Leukocyte Antigen \\
       11& CD45-KO & Cluster of Differentiation 45 - Antigen\\
        \bottomrule
\end{tabular}
\end{sc}
\end{scriptsize}
\end{center}
\vskip -0.1in
\end{table}

The synthesis of machine learning and biological domain knowledge holds potential for advancing scientific understanding and biomedical applications. While machine learning models excel at finding intricate patterns in data, their performance can be limited without the inductive biases, in particular for tabular data which typically lacks of it, and constraints derived from prior biological principles.

\subsection{Hierarchical Metrics Definition and Graph-Construction}
In our experiments, we construct the $k$-nearest neighbors graphs with $k=7$ neighbors. 
We perform a 7-fold test procedure where 1/7th of the data (2 or 3 patients out of 19) is held out as a test set and the rest (6/7th of data) as a training set. We then perform a 4-fold cross validation procedure within the training set to set the hyperparameters. 

The metrics are reported  using the extensions of the renowned metrics of precision, recall and F-score, but tailored to the HC setup. 
We implement the metrics of hierarchical precision ($hp$), hierarchical recall ($hr$) and hierarchical F-score ($hf$) defined by \cite{Kiritchenko2006}:
\begin{equation}
\scalebox{1.37}{$
    hp=\frac{\sum_i |\alpha_i \cap \beta_i|}{\sum_i |\alpha_i|}, ~hr=\frac{\sum_i |\alpha_i \cap \beta_i|}{\sum_i |\beta_i|},~ hf=\frac{2\times hp\times hr}{hp+ hr}.
$}
\end{equation}
where $\alpha_i$ is the set consisting of the most specific classes predicted for test sample $i$ and all their ancestor classes and $\beta_i$ is the set containing the true most specific classes of test sample $i$ and all their ancestors.  

\section{Interpretability and Feature Importance Analysis}
\label{appendix:feat-importance}

\begin{figure}[H]
    \centering
\includegraphics[width=0.65\textwidth]{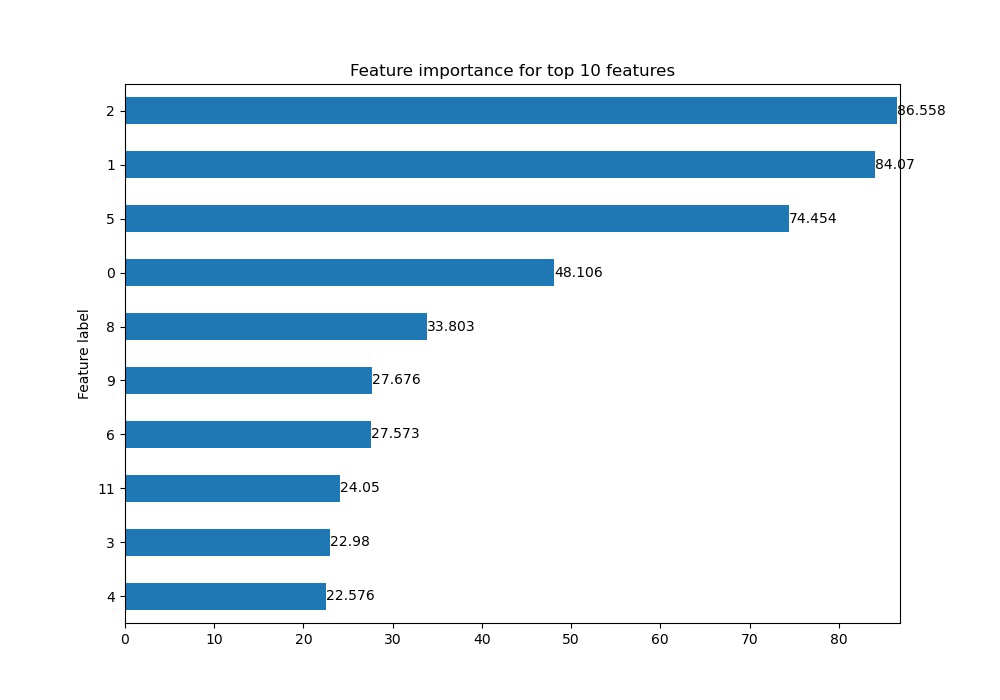}
    \caption{Feature importance for FCHC-GAT attributed by $\mathcal{H}$. Feature labels correspond to that of Table \ref{tab:1}.}
    \label{fig:features}
\end{figure}

An important aspect of our FCHC plug-in module is its interpretability, as demonstrated in Figure \ref{fig:features} for the FCHC-GAT version. 
This feature importance analysis not only provides insights into the decision-making process of our model but also paves the way for potential improvements and adaptations of our model for other applications.  As we can see, the features that are given the most importance (at least above 40\%) are 2,1,5 and 0, therefore we broke down the choice of the model for each one. In order of importance, we had:
\begin{itemize}
    \item CD14-FITC: This feature reflects the level of CD14 expression, a receptor found on the cell surface that interacts with lipopolysaccharide (LPS), a component of bacterial cell walls. Predominantly found on monocytes, macrophages, and activated granulocytes, CD14 plays a pivotal role in mediating the immune response to bacterial infections. The high weightage given to this feature suggests that our model is adept at distinguishing between cells involved in innate immunity and those that aren’t, thereby potentially detecting the presence of bacterial infection in the sample.
    \item Side Scatter (SSC)-Cell’s granularity: This feature measures the level of light scatter at a 90-degree angle to the laser beam, reflecting the internal complexity or granularity of the cell. This could include the presence of granules, nuclei, or other organelles. The high weightage of this feature implies that our model can differentiate between cells of varying complexity, such as lymphocytes, monocytes, and granulocytes, or identify cells with abnormal granularity, such as blast cells or malignant cells.
    \item CD33-PC5.5: This feature indicates the expression level of CD33, a cell surface receptor of the sialic acid-binding immunoglobulin-like lectin (Siglec) family. CD33 is expressed by myeloid cells like monocytes, macrophages, granulocytes, and mast cells, and it modulates the immune response by inhibiting the activation of these cells. The high weightage of this feature suggests that our model can distinguish between myeloid and non-myeloid cells, or detect the expression of CD33 as a marker for certain types of leukemia, such as acute myeloid leukemia (AML) or chronic myelomonocytic leukemia (CMML).
    \item Forward scatter (FSC)-Cell’s size: This feature measures the level of light scatter along the path of the laser beam, which is proportional to the diameter or surface area of the cell. This can be used to discriminate cells by size. The importance placed on this feature suggests that our model can differentiate between cells of different sizes, such as small lymphocytes and large monocytes.
\end{itemize}

\section{Graphical Representation of Multi-Head Attention}
\label{appendix:gat-viz}
\begin{figure}[ht]
\vskip 0.2in
\begin{center}
\centerline{\includegraphics[width=0.8\columnwidth]{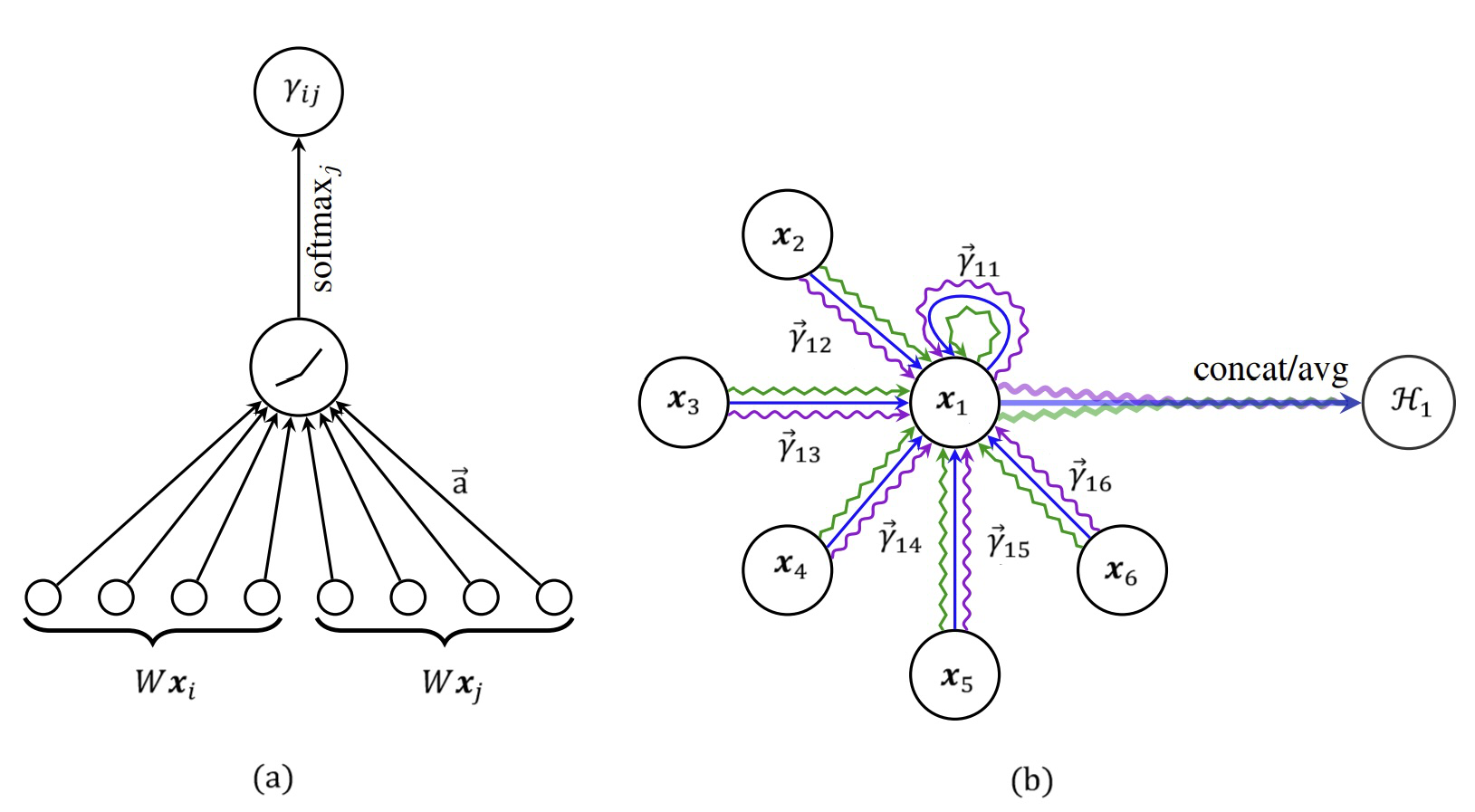}}
\caption{(a): Computing the normalized attention coefficients $\gamma_{ij}$  (b): Multi-head attention of node 1 on its neighborhood. Arrows show concatenation or averaging of attention (adapted from \protect\cite{Velickovic2017}).}
\label{fig:network}
\end{center}
\vskip -0.2in
\end{figure}

Figure \ref{fig:network} works as recap of the structural representation for the applied multi-head attention.
\newpage

\section{ImageNet1000-Mini Results}
\label{appendix:imagenet}
Table \ref{tab:imagenet1000res} shows a competitive performance of our FCHC-GNN module on hierarchical image classification task as well. For the sake of simplicity, we only showed our plug-in module being applied to a GAT, against its normal version and the average state-of-art performances (according to what was written on the benchmark \url{https://paperswithcode.com/sota/image-classification-on-imagenet}).
\begin{table}[ht]
\caption{Average accuracy percentage of SOTA methods on Imagenet1000-(Mini). Results are averaged with a $\text{std} = \pm 0.01$ across four different seeds. 
}
    \label{tab:imagenet1000res}
    \vskip 0.15in
\begin{center}
\begin{footnotesize}
\begin{sc}
\begin{tabular}{lcccr}
\toprule
        Model  & Imagenet1000-Mini\\
        \midrule
        \textbf{FCHC-GAT (Ours)}  & 91.88\\
        GAT  & 3.91 \\
         Average SOTA & 88.50 - \textbf{92.40} \\
\bottomrule
\end{tabular}
\end{sc}
\end{footnotesize}
\end{center}
\vskip -0.1in
\end{table}

\section{Ablation Studies on a Shallow Experiments Setup}
\label{appendix:1}
\begin{figure}[ht]
\vskip 0.2in
\begin{center}
\centerline{\scalebox{0.8}{\begin{forest}
  for tree={
    edge={->, thick},
    l=1.5cm
  }
  [Root
    [\textcolor{mygreen}{T cells}]
    [\textcolor{mygreen}{B cells}]
    [\textcolor{mygreen}{Monocytes}]
    [\textcolor{mygreen}{Mast cells}]
    [\textcolor{blue}{HSPC}
      [\textcolor{mygreen}{Myeloid HSPC}]
      [\textcolor{mygreen}{Lymphoid HSPC}]
      [\textcolor{mygreen}{Other HSPC}]
    ]
  ]
\end{forest}}}
\caption{Depiction of the shallow FC setup.}
\label{fig:shallowtree}
\end{center}
\vskip -0.2in
\end{figure}
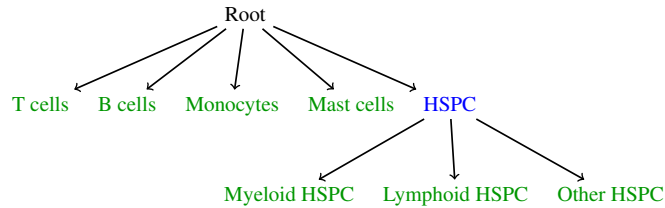

To compare the results given in Table \ref{tab:4} - \ref{tab:res-no-FCHCGNN}, we present the results on the second cohort of FC patients, provided by our Univerisity Hospitaly, where 30 different bone-marrows samples have been collected and anylized in the same way as the previous nineteens. The depth of hierarchy is shallower than that of the main one, where only one step of \textcolor{blue}{parental} nodes have been considered, as depicted in Figure \ref{fig:shallowtree}. As shown in Table \ref{tab:A2} and Table \ref{tab:A4} on shallow hierarchies, the problem more closely resembles a flat classification task, and thus baseline GNN performance remains relatively good. This is especially true for datasets with large populations of certain cell types, where the inherent class imbalance acts as an inductive bias, as depicted from the hierarchical F-Score results on Table \ref{tab:A3}. However, when the hierarchies grow deeper and more complex, the use of our FCHC-GNN module becomes strictly necessary to obtain better interpretability and generalization. For example, Table \ref{tab:A1} already shows a consistant improvement on the hF-Score metrics with respect to standard one without the hierarchical module, mainly due to the correct prediction of Myeloid HSPC and other minor HSPC cells. 

Deeper hierarchies exacerbate the issue of conflicting gradients between nodes at different hierarchy levels during training. Without explicit hierarchical constraints, as in the MCLoss, GNNs struggles to simultaneously optimize for accurate predictions at the \textcolor{mygreen}{leaf} node level while respecting \textcolor{blue}{parental} groupings (higher-level). Our hierarchical module resolves this tension without suffering of extra time-consumation problem as explained in Appendix \ref{appendix:time-complexity}, allowing attached models to leverage hierarchical priors as a strong inductive bias for learning more interpretable representations aligned with the known biological taxonomy.

\begin{table}[ht]
\caption{Average metrics across 30 patients using different models for FCHC module.}
 \label{tab:A1}
\vskip 0.15in
\begin{center}
\begin{small}
\begin{sc}
\begin{tabular}{lcccr}
\toprule 
        Metrics  & FCHC-GAT  
        & FCHC-SAGE & FCHC-GCN & FCHC-DNN \\
       \midrule
        hp  &  0.94
        & \textbf{0.97} & 0.95 &0.87 \\
        hr  & 0.93
        & \textbf{0.97} &  0.94& 0.83\\
        hf  &   0.94
        & \textbf{0.97} &0.94  & 0.85\\
\bottomrule
\end{tabular}
\end{sc}
\end{small}
\end{center}
\vskip -0.1in
\end{table}

\begin{table}[ht]
\caption{Average ratios of correct predicted classes across 30 patients using different models for FCHC module.}
    \label{tab:A2}
    \vskip 0.15in
\begin{center}
\begin{small}
\begin{sc}
\begin{tabular}{lcccr}
\toprule
        Label  & FCHC-GAT  
        & FCHC-SAGE & FCHC-GCN & FCHC-DNN\\
        \midrule
        T cells  &  93.87
        &\textbf{98.85}& 94.49& 75.04\\
        B cells  & 95.45
        &\textbf{98.28} &96.03 &86.82\\
        Monocytes  &  98.12
        &\textbf{99.21}& 98.57&95.24\\
        Mast cells &  99.79
        & \textbf{99.93}&99.79 & 99.79\\
        HSPC &  95.03
        & \textbf{97.85}& 95.73& 91.03\\
        Myeloid HSPC &  94.54 
        &\textbf{96.51}&94.79 &93.40\\
        Lymphoid HSPC &  97.40
        &\textbf{98.17}& 97.40& 97.40\\
        Other HSPC &  99.46 
        &99.46 &99.46 &99.46 \\

\bottomrule
\end{tabular}
\end{sc}
\end{small}
\end{center}
\vskip -0.1in
\end{table}

\begin{table}[ht]
\caption{Average metrics across 30 patients  without using the FCHC module.}
 \label{tab:A3}
\vskip 0.15in
\begin{center}
\begin{small}
\begin{sc}
\begin{tabular}{lcccr}
\toprule 
        Metrics  & GAT  
        & SAGE & GCN & DNN \\
       \midrule
        Precision  & 0.92 
        &0.84 & 0.92  & 0.84\\
        Recall  &  0.76
        &0.74 & \textbf{0.81}  & 0.71\\
        F1 Score  &  0.75
        &0.64 & \textbf{0.79}  & 0.61 \\
\bottomrule

\end{tabular}
\end{sc}
\end{small}
\end{center}
\vskip -0.1in
\end{table}

\begin{table}[ht]
\caption{Average ratios of corrected predicted classes across 30 patients without using the FCHC module.}
    \label{tab:A4}
    \vskip 0.15in
\begin{center}
\begin{small}
\begin{sc}
\begin{tabular}{lcccr}
\toprule
        Label  & GAT  
        & SAGE & GCN & DNN\\
        \midrule
        T cells  &  81.83
        &\textbf{98.72}& 90.14& 96.91\\
        B cells  &  51.50  
        &10.96&\textbf{57.22} & 9.87\\
        Monocytes  & \textbf{82.89} &  
         69.26& 80.89&56.19 \\
        Mast cells & \textbf{98.22} &  
      86.51  &52.62& 53.33 \\
        Myeloid HSPC &\textbf{26.16}  &  
       - &12.37&  -\\
        Lymphoid HSPC & \textbf{90.75}  &  
         12.96&84.50&   26.43 \\
        Other HSPC & 23.11  &  
       - &\textbf{32.48}&  - \\

\bottomrule
\end{tabular}
\end{sc}
\end{small}
\end{center}
\vskip -0.1in
\end{table}

\newpage
\section{FCHC Plug-In Module Implementation}
This section provides additional details on how the hierarchical constraint module was implemented in all the tested models. For sake of simplicity, we choose to illustrate how we designed the FCHC-GAT module, while for other FCHC-GNNs implementations refer to our GithHub repository \url{https://github.com/VIPER-GENEVA/FCHC-GNN-Hierarchical}. Here are the key steps:
\begin{itemize}
    \item It first defines the hierarchical structure by creating a directed graph $g$ from the class hierarchy string \textit{ATTRIBUTE\_class}, as shown in Listing \ref{code:MCM} below. This maps out the parent-child relationships between classes.
    \item It converts this graph to a matrix $R$ that encodes the ancestor-descendant relationships, where $R[i,j] = 1$ if class $i$ is descendant of class $j$, $0$ otherwise.
    \item Then the model includes GAT layers, where each of them does message passing and computes attention coefficients between neighboring nodes.
    \item During training, the raw GAT predictions are returned.
    \item During inference, the predictions are passed through a constraint function \textit{get\_constr\_out} that takes the max over descendants to enforce the hierarchy, as depicted in Listing \ref{code:leukograph} below.
    \item The loss function is a modified cross-entropy loss MCLoss that also lets subclasses inform about their parent classes predictions.
    \item The model takes a graph and node features as input, does message passing and attention in the GAT layers, applies the hierarchical constraint, and outputs a prediction vector for each node.
\end{itemize} In summary, it implements a hierarchical classifier using a GAT augmented with constraints and loss terms to incorporate the class hierarchy information. The hierarchy structure is predefined and encoded in the constraint matrix $R$.

\subsection{Constraint Layer}
The hierarchy consistency is imposed through a constraint layer added on top of the base classifier network. This constraint layer, referred to as the Max Constraint Module (MCM), takes the output predictions from the base network and ensures the hierarchical constraint is satisfied. The MCM output for any class \(A\) is computed as:
\begin{align}
    \text{MCM}_A = \max(\mathcal{H}_B \text{ where } B \text{ is subclass of } A) ,
\end{align} where $\mathcal{H}_B$ is the prediction value from the base network for class $B$. This takes the maximum over all subclasses of $A$, guaranteeing that the prediction for class $A$ will be greater than or equal to the prediction for any of its subclasses. By constructing the output this way, the hierarchy constraint is inherently satisfied regardless of the threshold used downstream.
\begin{figure}[ht!]
\centering
\begin{lstlisting}[caption={Python code snippet generating a matrix of ancestors $R$ based on the class relationships in the directed graph. Each class is represented as a node, and edges define parent-child connections.}, label={code:MCM}]
# List all the classes
to_skip = ['root']
ATTRIBUTE_class = "1,1_1,1_1_1,1_1_1_1,1_1_1_2,1_1_2,1_1_3,1_1_3_1,1_1_3_2,1_2,1_3,2"

# Store nodes and direct connections
g = nx.DiGraph()
for branch in ATTRIBUTE_class.split(','):
    term = branch.split('_')
    if len(term)==1:
        g.add_edge(term[0], 'root')
    else:
        for i in range(2, len(term) + 1):
            g.add_edge('.'.join(term[:i]), '.'.join(term[:i-1]))

nodes = sorted(g.nodes(), key=lambda x: (len(x.split('.')),x))

AA = np.array(nx.to_numpy_array(g, nodelist=nodes))

# Compute matrix of ancestors R
# Given C classes, R is an (C x C) matrix 
# where R_ij = 1 if class i is descendant of class j
R = np.zeros(AA.shape)
np.fill_diagonal(R, 1)
gg = nx.DiGraph(AA) # AA is the matrix where the direct connections are stored 
for i in range(len(AA)):
    ancestors = list(nx.descendants(gg, i)) # Need to use the function nx.descendants() 
    #because in the directed graph the edges have source 
    #from the descendant and point towards the ancestor 
    if ancestors:
        R[i, ancestors] = 1
R = torch.tensor(R)
#Transpose to get the descendants for each node 
R = R.transpose(1, 0)
R = R.unsqueeze(0).to(device)
\end{lstlisting}
\end{figure}

\subsection{Constrained Loss Function}
The loss function used during training is the max constraint loss (MCLoss), defined as:
\begin{align}
    \text{MCLoss}_A = -y_A \log\left(\max_{B \text{ in subclasses\_of\_A}} (y_B \cdot \mathcal{H}_B)\right) - (1 - y_A) \log(1 - \text{MCM}_A) ,
\end{align} where $y_A$ is the ground truth label. This loss allows the model to leverage the hierarchical information by letting predictions from subclasses inform the training for parent classes.

\subsection{Time Complexity}
\label{appendix:time-complexity}
The time complexity of the constraint layer is $O(C)$ where $C$ is the number of classes, since it just takes a maximum over the subclasses for each class. So it scales linearly with the size of the hierarchy and does not depend on the depth. The overall model maintains the same asymptotic time complexity as the base classification network used, with just an additional linear factor for the constraint layer. Hence, the hierarchical structure does not add significant overhead during training or inference.

\newpage
\begin{figure}[ht!]
\centering
\begin{lstlisting}[caption={FCHC-GAT plug-in module definition with hierarchical constraints. The model incorporates the hierarchy matrix $R$ to enforce class relationships during training. After performing message passing and attention computations, the model's output is constrained during inference using the \textit{get\_constr\_out} function.}, label={code:leukograph}]
def get_constr_out(x, R):
    ''' Given the output of the graph neural network x returns the output of 
    MCM given the hierarchy constraint expressed in the matrix R '''
    c_out = x.double()
    c_out = c_out.unsqueeze(1)
    c_out = c_out.expand(len(x),R.shape[1], R.shape[1])
    R_batch = R.expand(len(x),R.shape[1], R.shape[1])
    final_out, _ = torch.max(R_batch*c_out.double(), dim = 2)
    return final_out

# Define the FCHC plug-in module
class FCHC_GAT(nn.Module):
    ''' During training it returns the not-constrained output that is then passed to MCLoss'''
    def __init__(self,R):
        super(FCHC_GAT, self).__init__()
        self.R = R
        self.nb_layers = 2
        self.num_heads = 2
        self.out_head = 2
        self.hidden_dim = 32
        self.input_dim = 12
        self.output_dim= len(set(ATTRIBUTE_class.split(',')))+1  # We do not evaluate 
        #the performance of the model on the 'root' node
        gat_layers = []
        for i in range(self.nb_layers):
            
            if i == 0:                
                gat_layers.append(GATLayer(self.input_dim, self.hidden_dim, 
                self.num_heads,True,0.4))                
            elif i == self.nb_layers - 1:             
                gat_layers.append(GATLayer(self.hidden_dim*self.num_heads, 
                self.output_dim, self.out_head, False,0.2))               
            else:                
                gat_layers.append(GATLayer(self.hidden_dim*self.num_heads, 
                self.hidden_dim, self.num_heads, True,0.2))                
        self.gat_layers = nn.ModuleList(gat_layers)

        self.sigmoid = nn.Sigmoid()
        self.f = nn.ReLU()
        self.reset_parameters()  
        self.drop = nn.Dropout(0.2)
        
    def reset_parameters(self):
        for gat_layer in self.gat_layers:
            gat_layer.reset_parameters()
    
    def forward(self, data):
        x, edge_index= data.x, data.edge_index
        for i in range(self.nb_layers):
            x = self.gat_layers[i](x, edge_index)
            if i != self.nb_layers - 1:
                x = self.f(x)
                x = self.drop(x)
            else:
                x= self.sigmoid(x)
        if self.training:
            constrained_out = x
        else:
            constrained_out = get_constr_out(x, self.R )  
        return constrained_out
\end{lstlisting}
\end{figure}

\newpage
\section{t-SNE Embedding Visualization of FCHC-GAT Module}
\begin{figure}[ht]
\vskip 0.2in
\begin{center}
\centerline{\includegraphics[width=0.8\columnwidth]{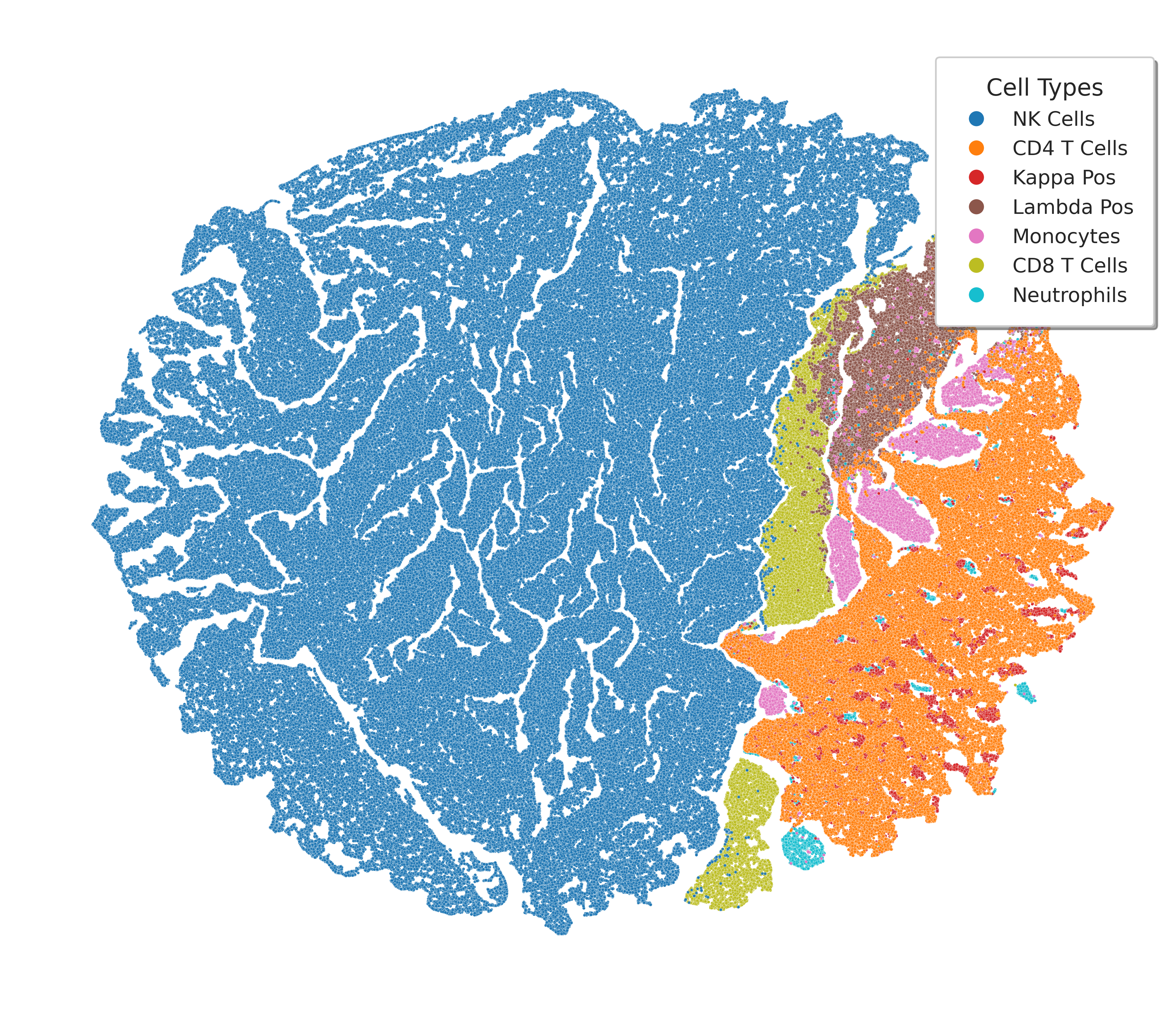}}
\caption{Embeddings t-SNE projection for patient 7, for the FCHC-GAT module.} 
\label{fig:tSNE}
\end{center}
\vskip -0.2in
\end{figure}

An important aspect of our FCHC-GAT model that warrants special attention is its ability to accurately capture and classify even small populations of certain cell classes. This is a significant achievement, as demonstrates the recall and precision of our model. The ability of FCHC-GAT is further illustrated through visualization of a particular sub-population of cells, particularly interested for physicians. As shown in  Figure \ref{fig:tSNE}, the t-SNE scatter plots provides a visual testament to the model’s ability to accurately classify different cell types of interested, in this case NK Cells, CD4 T Cells, Kappa Pos, Lambda Pos, Monocytes, Neutrophils and CD8 T cells.

\label{appendix:tsne}

\end{document}